%% file: main.tex
\documentclass[10pt,twocolumn,letterpaper]{article}

\usepackage[pagenumbers]{cvpr} 

\usepackage{graphicx}
\usepackage{amsmath}
\usepackage{amssymb}
\usepackage{booktabs}

\usepackage{bm}
\usepackage{multirow}
\usepackage{algorithm}
\usepackage[noend]{algpseudocode}

\usepackage[table]{xcolor}
\definecolor{LightCyan}{rgb}{0.88,1,1}
\definecolor{Gray}{gray}{0.8999}
\definecolor{LightGray}{gray}{0.9486}


\definecolor{cvprblue}{rgb}{0.21,0.49,0.74}
\usepackage[pagebackref,breaklinks,colorlinks,citecolor=cvprblue]{hyperref}

\def\paperID{11688} 
\def\confName{CVPR}
\def\confYear{2024}

\title{Masked AutoDecoder is Effective Multi-Task Vision Generalist}

\author{Han Qiu$^1$, Jiaxing Huang$^1$, Peng Gao$^2$, Lewei Lu$^3$, Xiaoqin Zhang$^4$, Shijian Lu$^{1}$\thanks{Corresponding author.}\\
\\
$^1$S-Lab, Nanyang Technological University \\
$^2$Shanghai Artificial Intelligence Laboratory, $^3$Sensetime Research\\
$^4$College of Computer Science and Technology, Zhejiang University of Technology\\
}

\begin{document}
\maketitle
\input{sec/0_abstract}    
\input{sec/1_intro}
\input{sec/2_relate}
\input{sec/3_method}
\input{sec/4_experi}
\input{sec/5_conclu}
{
    \small
    \bibliographystyle{ieeenat_fullname}
    \bibliography{main}
}

\clearpage
\appendix
\input{sec/X_suppl}

\end{document}

%% file: sec/0_abstract.tex
\begin{abstract}
Inspired by the success of general-purpose models in NLP, recent studies attempt to unify different vision tasks in the same sequence format and employ autoregressive Transformers for sequence prediction. They apply uni-directional attention to capture sequential dependencies and generate task sequences recursively. However, such autoregressive Transformers may not fit vision tasks well, as vision task sequences usually lack the sequential dependencies typically observed in natural languages. In this work, we design Masked AutoDecoder~(MAD), an effective multi-task vision generalist. MAD consists of two core designs. First, we develop a parallel decoding framework that introduces bi-directional attention to capture contextual dependencies comprehensively and decode vision task sequences in parallel. Second, we design a masked sequence modeling approach that learns rich task contexts by masking and reconstructing task sequences. In this way, MAD handles all the tasks by a single network branch and a simple cross-entropy loss with minimal task-specific designs. Extensive experiments demonstrate the great potential of MAD as a new paradigm for unifying various vision tasks. MAD achieves superior performance and inference efficiency compared to autoregressive counterparts while obtaining competitive accuracy with task-specific models. Code will be released at \url{https://github.com/hanqiu-hq/MAD}.
\end{abstract}

%% file: sec/1_intro.tex
\section{Introduction}
\label{sec:intro}

\begin{figure*}[t]
\centering
\includegraphics[width=0.98\linewidth]{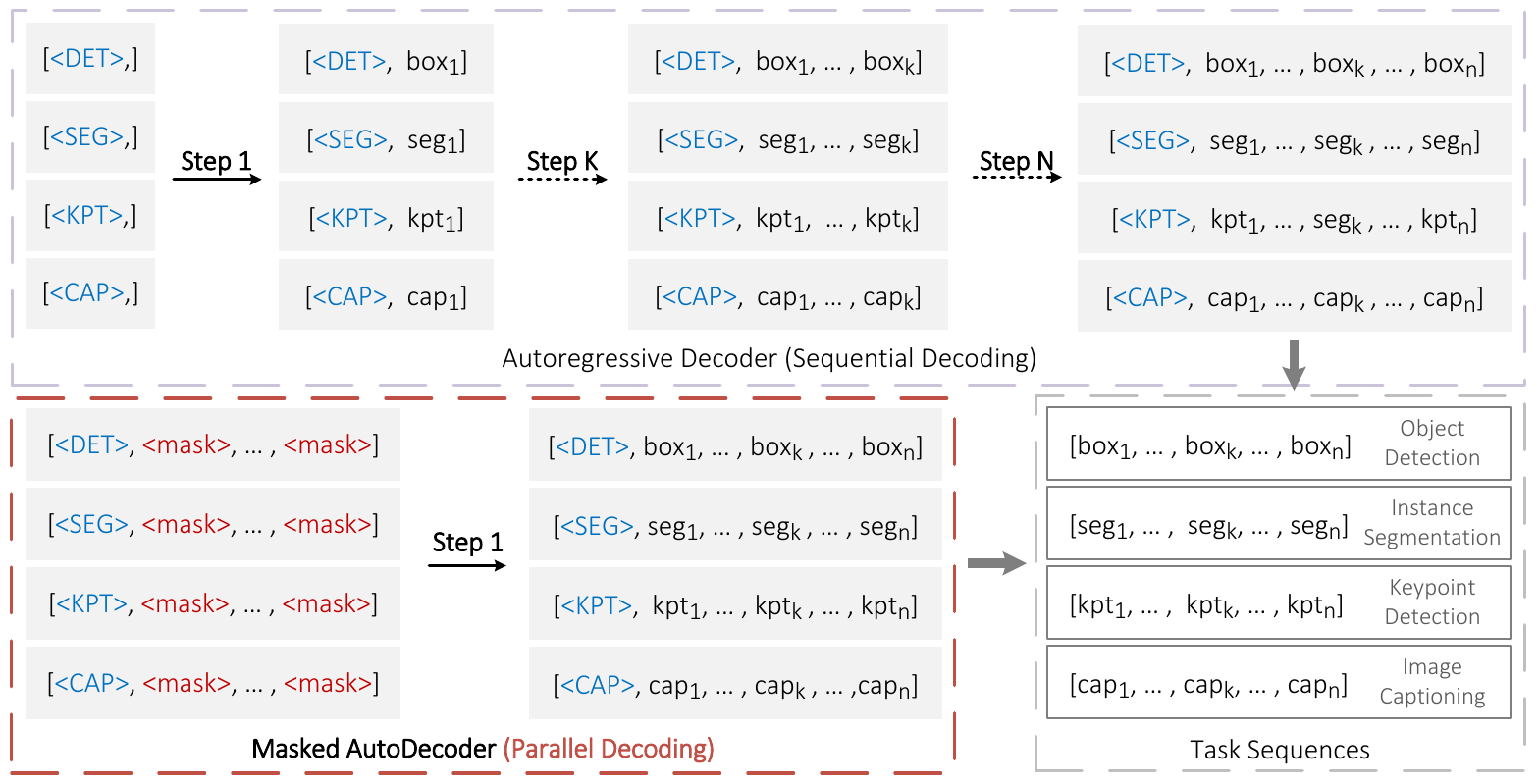}
\caption{Unified Sequence-to-sequence Modeling of Vision Tasks. The traditional \textit{Autoregressive Decoder} adopts sequential decoding for prediction, and utilizes unidirectional attention where each token can only attend to its previous ones. It generates task sequences token by token, resulting in a slow generation process with $N$ steps up to the length of sequences.  The proposed \textit{Masked AutoDecoder}~(MAD), equipped with parallel decoding with bidirectional attention and masked sequence modeling, allows decoding task sequences with only one step. Additionally, via masking and reconstructing task sequences, MAD can capture rich task contexts for different tasks, resulting in an effective and efficient vision generalist. Tokens in blue denote task prompts. $<mask>$ denotes mask token. \textit{Task Sequences} are simplified with details to be described in the ensuing Sections.}
\label{fig:intro}
\end{figure*}

Computer vision covers various concepts, such as localization, classification, and description, leading to a wide variety of highly structured outputs in different vision tasks, i.e., object detection, instance segmentation, keypoint detection, image captioning, etc. Following natural language processing~(NLP), recent methods~\citep{unified_io,pix2seqv2,wang2023visionllm,wang2022ofa,uvim} attempt to unify different vision tasks in an autoregressive sequence-to-sequence framework as illustrated in the upper part of Fig.~\ref{fig:intro}. They first model different vision tasks in the same sequence format, such as a sequence of coordinate and class label tokens for object detection, a sequence of contour coordinate tokens for image segmentation, or a sequence of descriptive sentences for image captioning. Additionally, the autoregressive Transformers~\citep{gpt3,radford2018improving,radford2019language}, with its specially designed uni-directional attention to capture sequential dependencies, are employed to recursively predict these vision task sequences.

\begin{figure}[t]
\centering
\includegraphics[width=\linewidth]{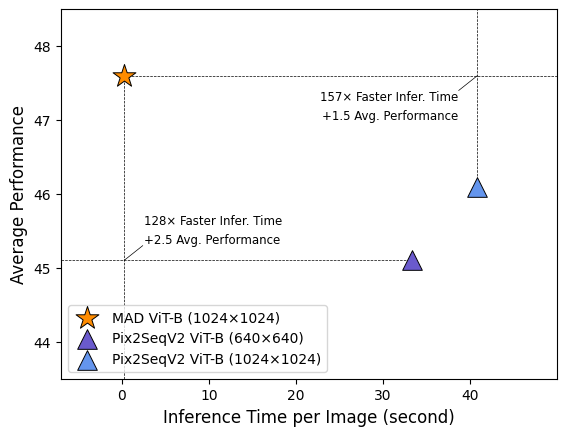}
\caption{The proposed MAD outperforms the state-of-the-art Pix2SeqV2~\citep{pix2seqv2} significantly in inference time, meanwhile achieves competitive accuracy across four representative vision tasks. The \textit{Average Performance} is averaged over four tasks including object detection~(mAP), instance segmentation~(mAP), keypoint detection~(mAP), and image captioning~(B@4). MAD achieves approximately 100$\times$ acceleration in inference time.}
\label{fig:teasor}
\end{figure}

Despite the success, the autoregressive approach often struggles on vision tasks due to two major factors: (1) \textit{The discrepancy between vision and language}. Language task sequences~\citep{gpt3,llama} heavily follow sequential dependencies while vision task sequences may not, e.g., the next word prediction in a sentence highly depends on its preceding texts, while the pixel prediction in segmentation tasks largely depends on its neighboring content instead of merely previous ones. The autoregressive approach, with uni-directional attention, can well capture sequential dependencies for language tasks but may not fit well with vision tasks. (2) \textit{Computation Efficiency}. The autoregressive approach predicts tokens in a sequence recursively, which is computation-intensive. The two factors might limit the model performance and efficiency, hindering the application of the autoregressive approach to vision tasks.

One possible solution for mitigating the above two issues is to explore bi-directional attention and parallel prediction for sequence modeling. This design leads to a customized Transformer that is capable of capturing more comprehensive dependencies and decodes the task sequence from scratch in parallel. However, such decoding process from scratch may struggle while modeling task contexts as sequences from different tasks highly vary in patterns, lengths, token vocabularies, etc., which will impede network convergence and result in inferior performance for multi-task learning.

Driven by the above analysis, we present Masked Decoder~(MAD), an effective sequence-based generalist for vision tasks. As illustrated in the bottom-left part of~\cref{fig:intro}, MAD masks tokens randomly from the task sequences and reconstructs the masked tokens based on the unmasked ones and image features, which provides rich task contexts for modeling disparate task sequences. In addition, it adopts an encoder-decoder Transformer architecture with bi-directional attention that leverages comprehensive dependencies in vision tasks to effectively decode task sequences in parallel. These designs enable a more efficient and effective multi-task learning framework that performs multiple vision tasks in a single architecture. Our experiments with four tasks~(object detection, instance segmentation. key-point detection, and image captioning) on the COCO dataset demonstrate that a simple MAD can achieve competitive accuracy and efficiency compared to both task-customized approaches and existing generalist models.

%% file: sec/2_relate.tex
\section{Related Works}
\label{sec:related_works}

\noindent\textbf{Vision Generalist Models} Learning a vision generalist model capable of handling multiple vision tasks using a shared architecture has long been a goal in computer vision. Inspired by the success of unified sequence-to-sequence based transformer framework~\citep{bert,radford2018improving, radford2019language} in natural language processing~(NLP), recent works~\citep{alayrac2022flamingo, wang2022ofa,reed2022generalist,chen2022pali} extend this framework to the field of computer vision and model various vision tasks in a unified sequence-to-sequence autoregressive paradigm. The pioneering works~\citep{pmlr-v139-cho21a, li2022blip,zhu2022uni} mainly focus on high-level semantic tasks, such as image captioning, visual question answering, image-text matching, and etc., considering their intrinsic correlation with language. In pursuit of unifying more vision tasks, especially for those involving dense predictions, Pix2seq~\citep{pix2seq,pix2seqv2} and UniTab~\citep{yang2022unitab} discrete object positions as a series of coordinate tokens to enable the localization capability for generalist models. Unified-IO~\citep{unified_io} and UViM~\citep{uvim} encode the per-pixel targets into semantic tokens for vision tasks that require outputs as images, such as depth estimation or panoptic segmentation. Uni-PercieverV2~\citep{li2023uni} equips an additional region proposal network to generate sequence predictions for object detection and instance segmentation. VisionLLM~\citep{wang2023visionllm} leverages LLM to enable flexible task output formats. Different from these methods which focus on customizing and extending more vision tasks in a sequence-based autoregressive framework, we demonstrate that such a framework may not fit well for vision tasks. Our masked Auto-decoding pursues a conceptually different direction and learns diverse task contexts in parallel via masked sequence modeling, leading to a more efficient and effective vision generalist.

\noindent\textbf{Masked Signal Modeling}
The paradigm of learning rich representations via masking and reconstructing has been widely explored in both fields of NLP and computer vision. In NLP, through masking and recovering language sentences, models like BERT~\citep{bert} and its variants~\citep{liu2019roberta,lan2019albert} successfully pre-train models capable of generalizing to a broad range of NLP tasks. In computer vision, such a paradigm also leads to multiple masked image modeling~(MIM)~\citep{convmae,bootMAE} and masked video modeling~(MVM) techniques. For example, BEIT~\citep{BEiT} explores MIM by recovering the masked image into visual tokens from discrete VAE~\citep{vae}. SimMIM~\citep{SimMIM}, MaskFeat~\citep{MaskFeat}, and MAE~\citep{MAE} incorporate low-level visual signals, such as RGB pixel value or the feature descriptor HOG~\citep{hog}, as the reconstruction targets. VideoMAE~\citep{videomae} encodes the corrupted video and learns to recover both spatial and temporal signals. The above methods employ masked signal modeling as a self-supervised task, aiming to learn to auto-encode rich representations for downstream tasks. Different from them, we propose masked auto-decoder~(MAD), exploring masked sequence modeling for decoding task sequences from its masked variants. Our approach is close to non-autoregressive translation~\citep{ghazvininejad2019mask,gu2017non} in NLP, but it has very different intrinsic objectives - non-autoregressive translation exploits parallel decoding to improve translation efficiency, while MAD aims to model diverse task contexts for learning multi-task vision generalists.

%% file: sec/3_method.tex
\section{Methods}
\label{sec:methods}

Our proposed unified generalist framework consists of three key components: (1) Unified tokenization of diverse input and output sequences for different tasks; (2) Masked auto-decoding framework for modeling task contexts; (3) An architecture that decodes desired task sequences based on image features. We introduce these components in the following sections.

\subsection{Task Tokenization}
\label{sec:tokenization}

In this work, we consider four vision-related tasks, including object detection, instance segmentation, keypoint detection, and image captioning. These tasks require the model's ability from classification to localization, from vision to language, and from image-level to pixel-level recognition. A comprehensive vocabulary is essential for dealing with such sophisticated problems. We build a universal vocabulary for all involved vision tasks and construct task sequences as follows:

For object detection, we convert bounding boxes into a sequence of tokens consisting of discrete coordinates and categories by the order of $[x_{min}, y_{min}, x_{max}, y_{max}, class]$. For implementation, we first construct a sequence consisting of $N$ noise objects, and then randomly replace and inject ground truth objects in the sequence. The $<Detection>$ prompt tokens are added before the sequence to identify the task. We set $N$ at 100 by default.

For instance segmentation, we directly predict the pixel mask following Mask R-CNN~\citep{he2017mask}. The bit masks of the size $M \times M$ are flattened and transformed into sequences consisting of $<Foreground>$ tokens and $<Background>$ tokens. We concatenate the prompt sequence consisting of task token $<Segmentation>$, bounding box coordinate tokens, and a class token to identify different instances.

For keypoint detection, we predict the coordinates and visibility for each keypoint of the person instance. It can thus be represented as a sequence of $[x, y, visibility, x, y, visibility, ...]$. We adopt two tokens $<Visible>$ and $<Invisible>$ to depict the visibility. The keypoints are arranged by the default order as in COCO dataset~\citep{coco}. For the occluded keypoints, we replace their coordinate tokens with random coordinates within the bounding box. We utilize the sequence $[<Keypoint>, x_{min}, y_{min}, x_{max}, y_{max}, person]$ to prompt keypoint detection task, where the coordinates in the prompt indicate the bounding box of the corresponding person.

For captioning, we adopt a pre-trained sentence-piece model~(SPM)~\citep{kudo2018sentencepiece} to convert a caption into a sequence of discrete tokens. We randomly replace one of the tokens in the transferred sequence with a random word token for sequence augmentation. All the sequences are padded or truncated to a length of 20 tokens. The $<Caption>$ token is adopted as the prompt.

The final vocabulary thus comprises five parts, including prompt tokens for distinguishing tasks, coordinate tokens for localization, category tokens for classification, task-related special tokens, and word tokens for captioning. Compared to previous studies~\citep{pix2seqv2,unified_io,wang2023visionllm}, we modify the format of task sequences accordingly in MAD for embracing parallel decoding.


\begin{figure*}[t]
\centering
\includegraphics[width=0.95\linewidth]{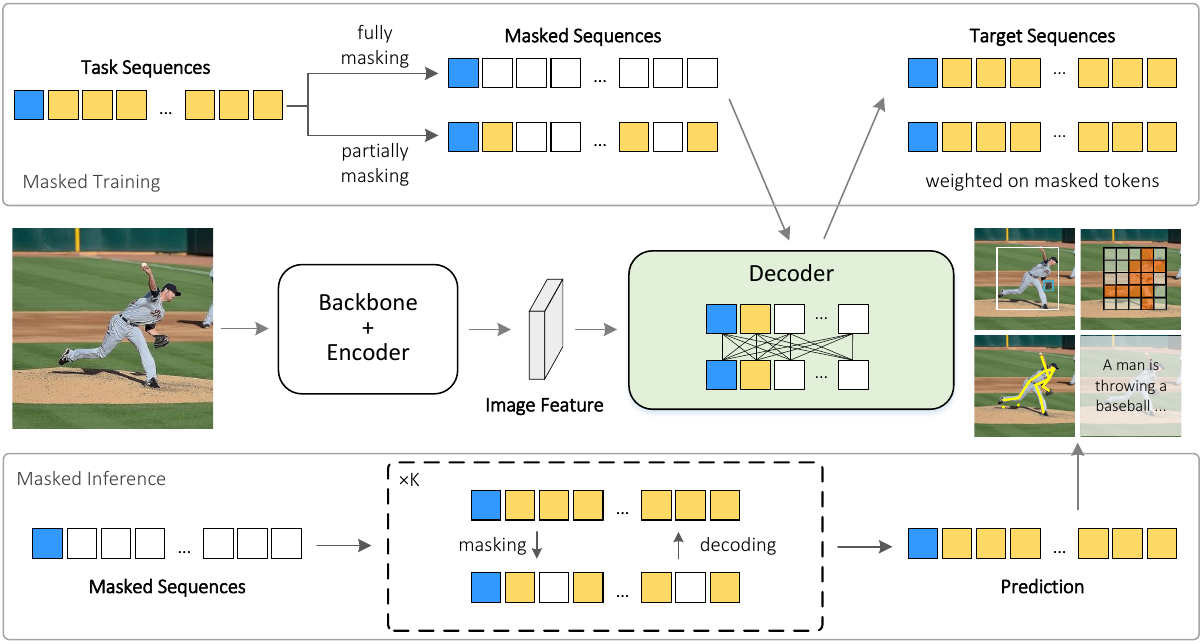}
\caption{Illustration of our proposed MAD with masked training and masked inference. MAD consists of two major parts, a \textit{Backbone + Encoder} to extract the representation of the input images and a \textit{Decoder} that processes  \textit{Masked Sequences} for prediction. During training, MAD randomly masks \textit{Task Sequences}~(blue for prompt tokens and yellow for task tokens) to generate \textit{Masked Sequences}~(white for masked tokens) and learns to reconstruct the \textit{Task Sequences}. During inference, MAD takes fully \textit{Masked Sequences} as input and repeats the decoding and masking process to refine predictions. When \textit{K} in masked inference is set to 0, MAD can skip refinement and directly generate prediction in one step.}
\label{fig:architecture}
\end{figure*}

\subsection{Masked AutoDecoding}
\label{sec:masking}
\textbf{Masked Training} We propose Masked AutoDecoding for multi-task sequence modeling. We randomly sample a subset of target tokens and mask the remaining ones. The sampling follows a uniform distribution. The masked tokens are replaced by special $<Mask>$ tokens, which are shared among all tasks. During training, we adopt two kinds of sequences for each task, a fully masked sequence and a partly masked sequence as shown in~\cref{fig:architecture}.

The reconstruction of fully masked sequences establishes a basis to train a unified decoder, which is, learning to decode multi-task sequences with only the prompts. Therefore, all the tokens, except those in the prompt sequences, are replaced with $<Mask>$ before being fed into the decoder. The training objective is to reconstruct the desired task sequences based on task prompts. However, different from the autoregressive approach where each task sequence is specified by its corresponding input sequence, a fully masked sequence in auto-decoding might match multiple similar task sequences, such as differently arranged objects for object detection or similar captioning sentences per image for image captioning. Randomly choosing the reconstruction target each time might hinder convergence. Hence, we adopt \textit{Hungarian Matching}~\citep{kuhn1955hungarian} to construct the task sequences for object detection and image captioning. For instance segmentation and keypoint detection, the original unmasked sequences are adopted as targets since their prompt sequences with object locations and categories are able to specify the unambiguous task sequences.

However, when all the tokens are masked, it is difficult for the model to distinguish different task sequences based on only a few prompt tokens. We thus leverage partly masked sequences to alleviate this issue. The unmasked tokens provide rich cues for the pattern of different task sequences, which help the decoder capture diverse task contexts. During training, both fully and partly masked sequences are concatenated together and decoded in parallel.

The MAD task is greatly inspired by the self-supervised masked auto-encoding approach in both language and vision domains, which learns and encodes informative representation by reconstructing masked content. We expand this idea of masked modeling to decode multi-task sequences in computer vision. This simple method, by modeling corrected sequences and predicting missing tokens, enables MAD to learn distinct task contexts and inter-sequence dependencies for vision tasks.

\noindent\textbf{Masked Inference} During inference, we conduct multi-stage masked decoding to refine the prediction. With the initial prediction recovered from the fully masked sequence, we randomly sample tokens from the predicted sequences and replace them with mask tokens. The corrupted sequences are then fed to the decoder for reconstruction. We directly ensemble predictions from masked tokens to their original tokens to obtain more accurate results.

\subsection{Architecture}
Our goal is to build a single model that is capable of handling different vision-related tasks within a unified sequence paradigm with little task-customized designs. Hence, we adopt a simple encoder-decoder transformer architecture, which has been proven successful in handling sequences with variable lengths in both natural language processing and computer vision tasks. As shown in Fig.~\ref{fig:architecture}, the overall architecture of MAD consists of two parts, a backbone network with the encoder to extract image features and a decoder to reconstruct the masked sequences. 

\noindent\textbf{Backbone and Transformer Encoder.} Given an input image $I\in \mathbb{R}^{H\times W \times 3}$, a backbone network is adopted to generate a low-resolution image feature with a stride of 32. The encoder then takes the image feature, adds 2D positional encodings, and processes the feature via a series of encoder layers consisting of a self-attention module and feed-forward network~(FFN). The image feature is then injected into the decoder as a condition to decode task sequences.

\noindent\textbf{Transformer Decoder.} The decoder follows standard architecture, reconstructing \textit{Masked Sequences} through self-attention, cross-attention, and FFN layers. To address the sequence order, we introduce learned sequence positional encodings and add them to the input sequences before each attention layer in the decoder. The sequence positional encodings are shared among all the tasks and are truncated according to the length of different task sequences. Unlike existing autoregressive methods~\citep{pix2seqv2,unified_io} that adopt uni-directional masks in self-attention layers, our model decodes all the sequence in parallel with bi-directional attention, leading to more efficient and effective predictions and multi-task training.


\subsection{Multi-task Training}
\textbf{Loss Function.}
We adopt a softmax cross-entropy loss to maximize the likelihood of masked sequence conditioned on the image feature:

\begin{equation}
    L = \sum_{t} W_{t} \frac{1}{N_{m}}\sum_{i\in M}log P(\hat{y}_{i}|x, y)
\end{equation}

where $y$ and $\hat{y}$ are masked and decoded sequences, $W_{t}$ is loss weights for different tasks, $M$ means the set of masked tokens, and $N_{m}$ denotes the number of masked tokens. Only the loss of masked tokens is counted. Following previous practice~\citep{DETR,al2019character}, we adopt auxiliary losses for the predictions after each decoder layer. For each task, we filter the target token vocabulary so that losses are only calculated on its involved vocabulary to improve training 
efficiency. Considering tokens of the whole vocabulary leads to intensive computation and memory usage since image captioning involves plenteous text tokens that are not involved in other tasks.

\noindent\textbf{Task Mixed Sampling.} For learning a single model for multiple tasks, we employ a task-mixed sampling strategy where each image in the dataset is sampled with its annotations mixed from all tasks. The sampled images are processed by the backbone and encoder only once for encoding image features shared by all tasks. Only the decoding process is repeated for different tasks, considering that they hold different sequence lengths and are hard to process in parallel. Such a strategy is conceptually simple and effective compared with the batch mixing strategy from existing work~\citep{pix2seqv2,li2023uni} where each batch only samples image-sequence pairs for a single task. Considering that each image might involve multiple vision tasks, batch mixing requires encoding the same image multiple times for different tasks. As a comparison, task mixing provides a more flexible framework to add more data from more tasks, while also sharing most model components among tasks, resulting in better efficiency.

\subsection{Comparison with Autoregressive Decoding}
The central idea of MAD is conceptually simple: parallel decoding with bidirectional attention for effective and efficient task-sequence decoding, and masked autodecoding for task-context modeling. Although the second design shares a similar concept as the previous autoregressive counterpart by removing part of the data and learning to recover them, our approach is intrinsically different. Take an autoregressive vision generalist model Pix2SeqV2~\citep{pix2seqv2} as an example. The differences can be summarized in four major aspects as listed.

\noindent\textbf{Sequence Data Corruption Strategies.} The way that autoregressive approaches corrupt data is implicit, by applying the uni-directional mask on the attention layer to restrict tokens from perceiving their following ones. On the contrary, MAD achieves corruption by directly masking tokens of input sequences.

\noindent\textbf{Content Recovering Strategies.} Pix2SeqV2 adopts next-token prediction, where each token of sequences is trained to predict its subsequent token. On the contrary, MAD predicts masked tokens only with original tokens as targets.

\noindent\textbf{Self-Attention Mechanisms.} The triangular causal mask is applied to the self-attention layer of the decoder in Pix2SeqV2, resulting in uni-directional attention. On the contrary, MAD adopts bi-directional self-attention for the decoder, which facilitate in capturing comprehensive dependencies for vision task sequences.

\noindent\textbf{Inference Mechanisms.} Pix2SeqV2 leverages sequential decoding that generates sequences token by token, resulting in slow inference time that increases with sequence length. On the contrary, MAD adopts parallel decoding that could generate task sequences with only one step. In addition, MAD can further refine predictions by repeating the making and decoding process, leading to a more efficient and flexible inference mechanism.

%% file: sec/4_experi.tex
\section{Experiments}
\label{sec:experiments}

\subsection{Experimental Settings}

\begin{table*}[]
    \footnotesize
    \centering
        \caption{Comparisons for object detection~(mAP), instance segmentation~(mAP), keypoint detection~(mAP), and image captioning~(BLEU@4~\citep{papineni2002bleu}) on COCO validation set.}
    \begin{tabular}{lccccccc}
         \toprule
          & Backbone & Parameter(M) & Infer. Time(s) & Object Det. & Instance Seg. & Keypoint Det. & Captioning \\
         \midrule 
         \rowcolor{Gray} \multicolumn{8}{l}{\textbf{\textit{Task-specific Models}}} \\
         Faster R-CNN~\citep{ren2015faster} & R101-FPN & 42M & - & 42.0 & - & - & - \\
         DETR~\citep{DETR} & R101-DC5        & 60M & - & 44.9 & - & - & - \\
         Pix2Seq~\citep{pix2seq} & R101-DC5     & 57M & - & 45.0 & - & - & - \\
         Mask R-CNN~\citep{he2017mask} & X101-FPN  & 107M & - & 42.9 & 38.6 & - & - \\
         Keypoint R-CNN~\citep{detectron2} & R50-FPN & 59M & - & - & - & 65.5 & - \\
         Transformer~\citep{sharma2018conceptual} & Encoder & - & - & - & - & - & 34.0 \\
         \midrule
         \rowcolor{Gray} \multicolumn{8}{l}{\textbf{\textit{Generalist Models}}} \\
         VisionLLM~\citep{wang2023visionllm} & R50+Alpaca-7B & 40M + 7B & - & 44.6 & 25.1 & - & 31.0 \\
         Pix2SeqV2~\citep{pix2seqv2} & ViT-B & 132M & 40.86 & 46.5 & 38.2 & 64.8 & 34.9 \\
         \textbf{MAD} & Swin-B & 107M & 0.19 & 49.7 & 40.6 & 64.6 & 32.2 \\
         \textbf{MAD} & ViT-B & 107M & 0.26 & 50.1 & 41.2 & 66.5 & 32.6 \\
         \bottomrule
    \end{tabular}
    \label{tab:main_result}
\vspace{-10pt}
\end{table*}

\noindent\textbf{Dataset and Tasks.}
Following previous practice~\citep{pix2seqv2,wang2023visionllm}, we evaluate MAD on MS-COCO dataset~\citep{coco} which contains 118k training images and 5k validation images with annotations for all four tasks we considered. For object detection, we take $N=100$ instances per image for training, resulting in a sequence of length 500. The coordinates of bounding boxes are discretized into 500 bins. For instance segmentation, we randomly sample ten instances and transform their segmentation masks into bit masks with a size of $16 \times 16$. For keypoint detection, we train MAD on ten person instances per image and only predict keypoints for detected humans~(based on object detection results) during inference. We pad blank instances for the above three tasks if there are not enough instances in the image. For image captioning, we adopt sentence piece model~(SPM) from T5~\citep{t5} for tokenization, and abbreviate its vocabulary based on COCO dataset, resulting in 11421 remaining text tokens. We use loss weights of [1.5, 2.7, 0.5, 0.3] for object detection, instance segmentation, keypoint detection, and image captioning respectively.

At inference time, different tasks in MAD can be conducted in any order. During implementation, we first predict task sequences for object detection as they will serve as prompts for subsequent tasks. For instance segmentation, we directly convert predicted sequences into bit masks based on probabilities of $<Foreground>$ tokens. For keypoint detection, predicted sequences are dequantized into tuples of keypoint coordinates with probabilities of $<Visible>$ showing their visibility. For image captioning, sequences are truncated by the first predicted padding token and directly mapped back to texts by SPM. We conduct masked inference on keypoint detection and image captioning tasks with mask ratios of {0.7}, and {0.8, 0.6, 0.4} respectively.

\noindent\textbf{Implementation Details.} We implement MAD with three different backbones, Swin-Base~\citep{SwinTransformer}, ViT~\citep{ViT} and Resnet-50~\citep{resnet} with detrex~\citep{detrex} toolbox. For ViT, we adopt the adjusted architecture of ViTDet~\citep{vitdet}~(for single feature map output). Following DETR~\citep{DETR}, both the encoder and decoder in MAD consist of 6 layers with a main dimension of 256 and 8 attention heads, and the width of FFN is set to 2048. For sequence modeling, we adopt learned positional encodings with a length of 506 to cover all task sequences.

For comparisons with state-of-the-art methods, we train the model with Swin-Base~\citep{SwinTransformer} and ViT-Base~\citep{ViT,eva02} for 300 and 100 epochs with a learning rate drop after 200 and 80 epochs. Our ablation experiments with the ResNet-50 backbone are trained with a shorter schedule of 50 epochs. More training details can be found in the supplementary material. The inference time~(speeds) of all experiments are the total time for inferring on four tasks, tested on a single A100 with a batch size of one image.

\subsection{Comparison with State-of-the-art Methods}
\cref{tab:main_result} shows comparisons with the state-of-the-art~(SOTA). We compare MAD with two types of SOTA models: (1) typical task-specific models which leverage task-specific designs and are trained on a single task; (2) generalist models which employ a shared single architecture to handle multiple vision tasks without task-specific designs such as region proposal network~(RPN) or ROI Pooling. Compared with the task-specific models, we can see that MAD can achieve competitive and even better accuracy without customized architecture for a single task. On top of that, the sequence-based framework in MAD provides significant scalability and flexibility to new tasks or data formats than these models. We adopt a MIM pre-trained ViT from~\citep{eva02} for comparison with Pix2SeqV2 \citep{pix2seqv2} which is pre-trained on Objects365~\citep{objects365}. \cref{tab:main_result} shows that MAD outperforms Pix2SeqV2 by a large margin, especially on vision-centric tasks, demonstrating the effectiveness of our designs on bidirectional attention and masked auto-decoding strategy. Additionally, with the same backbone, MAD is around 157x faster than Pix2SeqV2 (with sequential decoding and complicated post-processing) in inference time. For image captioning, the autoregressive paradigm in existing methods excels us in modeling language sequential context. We will investigate how to combine the advantages of both in the future to enable a more versatile generalist model.

\begin{table}[]
    \footnotesize
    \centering
    \caption{Ablation studies on main components of MAD. ``AR Decoding'' indicates autoregressive decoding.}
    \begin{tabular}{lccccc}
         \toprule
         Methods & Time~(ms)  & Det. & Seg. & Kpt. & Cap. \\
         \midrule 
         AR. Decoding & 3953 & 27.9 & 12.3 & 33.4 & 34.1 \\
         + Parallel Decoding & 137 & 35.9 & 29.8 & 51.5 & 18.2 \\
         + Masked Training  & 137 & 38.9 & 32.3 & 54.6 & 18.6 \\
         + Masked Inference & 173 & 38.9 & 32.3 & 54.7 & 29.6 \\
         \bottomrule
    \end{tabular}
    \label{tab:main_components}
    \vspace{-5pt}
\end{table}

\begin{table}[t]
\centering
\footnotesize
\caption{Ablations on Masking ratio strategies. For ``random ratio'', we used a random mask ratio lying between 0.6 and 0.8. For ``multiple ratios'', the task sequences are trained with two masking ratios~(i.e., 0.6 and 0.8). The ``single ratio'' indicates that a single mask ratio is adopted.}
    \begin{tabular}{ccccc}
        \toprule
        Methods & Det. & Seg. & Kpt. & Cap. \\
        \midrule 
        random ratio & 38.6 & 31.9 & 54.3 & 29.4 \\
        multiple ratios & 38.6 & 32.0 & 54.2 & 29.7 \\
        \textbf{single ratio} & 38.9 & 32.3 & 54.7 & 29.6 \\
        \bottomrule
        \end{tabular}
    \label{tab:maskmethod}
\end{table}

\subsection{Ablation Studies}

\noindent\textbf{Main Components Ablation.} We first gradually ablate our main designs as shown in~\cref{tab:main_components}. We convert MAD into an autoregressive variant with the same architecture for comparison~(Details can be found in the supplementary material). It can be seen that \textit{Autoregressive Decoding} performs worst in terms of both inference time and accuracy on vision tasks except image captioning. This result is consistent with our analysis that the autoregressive approach might not fit well for vision-centric tasks and struggles with extremely slow predictions. By employing bi-directional attention and parallel decoding~(i.e., \textit{Parallel Decoding}), the convergence and inference speed of vision tasks are greatly improved. We then introduce masked autodecoding. As shown in the fourth row, \textit{+Masked training} performs especially better for object detection, instance segmentation, and keypoint detection, thanks to the task context modeled through masking and reconstruction. Moreover, by further introducing masked inference (i.e., \textit{+Masked Inference}), the accuracy is constantly improved with competitive image caption accuracy to the autoregressive counterpart. In addition, we observe that MAD has different effects on vision-centric tasks and language tasks in training and inference. We speculate that MAD in training could model rich task contexts, such as the relationship among task prompts, vocabulary, and sequence patterns, which are crucial for modeling multi-task sequences. On the contrary, during inference, MAD mainly focuses on dependencies among sequence tokens, which are generally rich in language but lacking in visual sequences.

\begin{figure*}[t]
    \centering
        \includegraphics[width=0.97\linewidth]{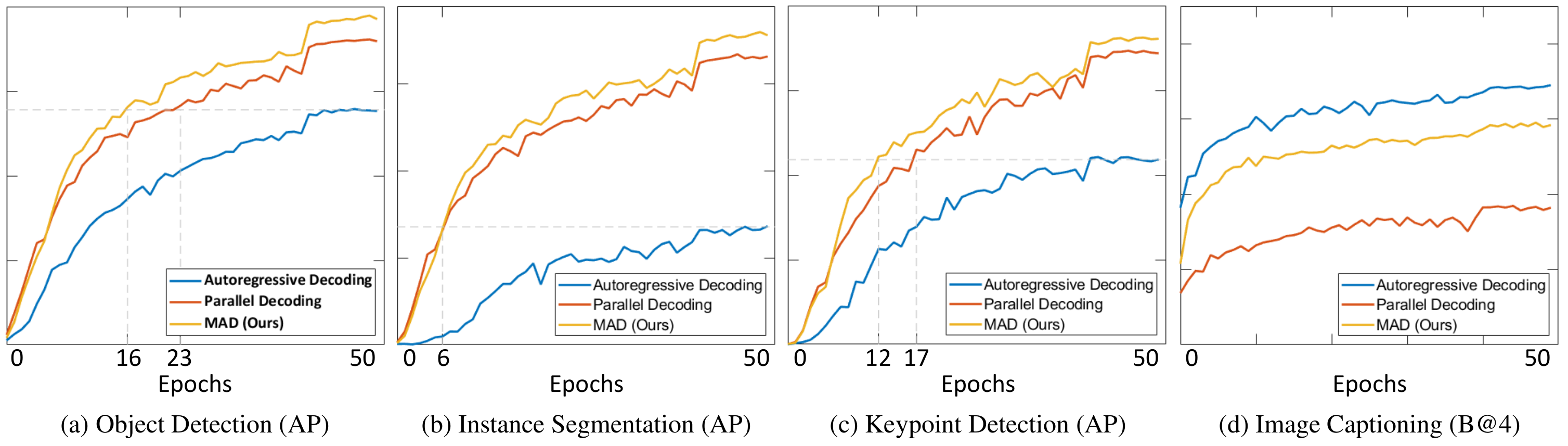}
    \caption{Convergence curves for \textit{Autoregressive Decoding}, \textit{Parallel Decoding}, and the proposed \textbf{MAD} in~\cref{tab:main_components}. MAD achieves much faster convergence for vision-centric tasks and greatly narrows the gap with \textit{Autoregressive Decoding} compared with \textit{Parallel Decoding} for image captioning.}
    \label{fig:convergence}
\end{figure*}

\noindent\textbf{Convergence Curves for Vision Tasks.}~\cref{fig:convergence} compares the detailed training curves between methods in~\cref{tab:main_components} for different tasks. With bi-directional attention, both MAD and parallel decoding converge much faster than autoregressive decoding which adopts uni-directional attention. In addition, the masked sequence modeling strategy in MAD can further capture rich task contexts and largely improve performances, especially for image captioning. These results further demonstrate the non-trivial design of MAD.

\begin{table}[t]
\centering
\footnotesize
\caption{Ablation on different mask ratios in training.}
    \begin{tabular}{ccccc}
         \toprule
         Mask Ratio & Det. & Seg. & Kpt. & Cap. \\
         \midrule 
         0.4 & 38.4 & 31.8 & 54.4 & 29.2 \\
         0.6 & 38.5 & 31.9 & 54.2 & 29.7 \\
         \textbf{0.7} & 38.9 & 32.3 & 54.7 & 29.6 \\
         0.8 & 38.7 & 32.1 & 54.9 & 28.4 \\
         \bottomrule
    \end{tabular}
    \label{tab:maskratio}
\end{table}

\begin{table}[t]
\centering
\footnotesize
\caption{Ablations on the number of quantization bins for coordinates.}
    \begin{tabular}{cccc}
         \toprule
         Number of Bins. & Object Detection & Keypoint Detection \\
         \midrule 
         300 & 38.5 & 54.2 \\
         \textbf{500} & 38.9 & 54.7 \\
         800 & 38.6 & 54.5 \\
         1000 & 38.5 & 54.4 \\
         \bottomrule
    \end{tabular}
    \label{tab:numofbin}
\end{table}

\noindent\textbf{Masked Training.} We examine how varying masking strategies and masking ratios affect the training of MAD. As~\cref{tab:maskmethod} shows, the simplest strategy with a single masking ratio could achieve the highest performance. As for specific masking ratios in training (under the \textit{single ratio} strategy), ~\cref{tab:maskratio} shows that MAD performs the best with a moderate value of 0.7, while a smaller masking ratio results in an over-simplified task, and a larger masking ratio leaves insufficient tokens for modeling task contexts.

\noindent\textbf{Coordinate Quantization.} We evaluate the effect of the number of the coordinate bins. As~\cref{tab:numofbin} shows, MAD performs robustly under different numbers of bins. We thus adopt 500 as default, while each bin corresponds to approximately 2 pixels for an image with a size between 800 to 1333 pixels, resulting in negligible quantization error.

\begin{table}[t]
\centering
\footnotesize
\caption{Ablations on size of bit mask for image segmentation.}
    \begin{tabular}{ccc}
         \toprule
         Mask Size & Object Detection & Instance Segmentation \\
         \midrule 
         12 & 38.4 & 31.5 \\
         14 & 38.8 & 32.0 \\
         \textbf{16} & 38.9 & 32.3 \\
         20 & 38.8 & 32.4 \\
         \bottomrule
    \end{tabular}
    \label{tab:masksize}
\end{table}

\noindent\textbf{Mask Size.} In~\cref{tab:masksize}, we study the size of the segmentation mask. It can be seen that MAD does not benefit much from larger mask sizes, since we do not adopt task-specific operations like ROIAlign~\citep{he2017mask} or interpolation to align mask pixels and image pixels. Considering that larger mask sizes lead to longer task sequences, we set the mask size at 16 for good efficiency.

\begin{table}[t]
\centering
\footnotesize
\caption{Ablations on inference mask ratios for image captioning.}
    \begin{tabular}{lc}
         \toprule
         Mask Ratio & Image Captioning \\
         \midrule 
         w/o masked inference & 18.6 \\
         \{0.7\} & 25.8 \\
         \{0.7, 0.3\} & 27.0 \\
         \textbf{\{0.8, 0.6, 0.4\}} & 29.6 \\
         \bottomrule
    \end{tabular}
\label{tab:infermask}
\end{table}

\noindent\textbf{Inference Mask Ratio for Captioning.} We examine different inference mask ratios for image captioning. Results in~\cref{tab:infermask} demonstrate that a combination of gradually decreasing masking ratios~(\{0.8, 0.6, 0.4\}) performs the best.

\subsection{Joint Training}
\noindent\cref{tab:joint_train} shows the performance of MAD under separate training~(single-task) and joint training~(multi-task), both being equipped with our proposed masked autodecoding. It can be observed that joint training obtains more gains with masked autodecoding. We conjecture that under separate training, the sequence encodings as well as other learnt weights in the decoder are capable of modeling task context of a single task. However, as more vision and language tasks are introduced simultaneously, the jointly trained model might fail to model diverse task contexts and therefore benefit more from our proposed masked autodecoding.

\begin{table}[t]
\centering
\footnotesize
\caption{Comparing joint training and separated training in MAD. \textit{MM} denotes the proposed masked autodecoding including both \textbf{M}asked training and \textbf{M}asked inference. For \textit{separate training}, the model is trained on a single task. The models \textit{w/o MM} are conducted with only parallel decoding.}
    \begin{tabular}{lcccc}
         \toprule
         Methods & Det. & Seg. & Kpt. & Cap. \\
         \midrule 
         separate training (w/o MM) & 38.4 & 31.2 & 55.1 & 20.6 \\
         separate training (w/ MM)  & 38.5 & 32.1 & 55.7 & 30.4 \\
         \midrule
         joint training (w/o MM) & 35.9 & 29.8 & 51.5 & 18.2 \\
         joint training (w/ MM) & 38.9 & 32.3 & 54.7 & 29.6 \\
         \bottomrule
    \end{tabular}
    \label{tab:joint_train}
\end{table}

%% file: sec/5_conclu.tex
\section{Conclusion}

In this work, we propose Masked AutoDecoder~(MAD), a sequence-to-sequence multi-task vision generalist that employs masked sequence modeling and parallel decoding. MAD performs multiple vision tasks with a unified task sequence format, and learns to reconstruct masked task sequences for modeling diverse task contexts. In addition, we employ bidirectional attention and parallel decoding in Transformer, achieving significant speedup in both convergence and inference compared to autoregressive counterparts for vision tasks. Experiments on COCO demonstrate the effectiveness and superiority of MAD as compared with both well-established task-specific models and existing vision generalist models.

\noindent\textbf{Limitations and Future Work.} In this work, we only explore MAD on four typical vision tasks following~\cite{pix2seqv2}. We plan to expand MAD towards more data and tasks in the future work to form a more comprehensive vision generalist and further explore its characteristics.

\section*{Acknowledgement}
This study is supported under the RIE2020 Industry Alignment Fund – Industry Collaboration Projects (IAF-ICP) Funding Initiative, as well as cash and in-kind contribution from the industry partner(s).

%% file: sec/X_suppl.tex
\maketitlesupplementary

\section{Implementation Details}
~\cref{tab:exp_setting} shows the default settings of our experiments. Most of the hyper-parameter setting and training strategies follow DETR~\citep{DETR}. We adopt the AdamW optimizer with a learning rate of 1e-4 for all the experiments. The backbone is fine-tuned with a smaller learning rate of 1e-5. We use scale augmentation for ResNet-50 and Swin-Base models. Specifically, the shortest side of the input image is resized to between 480 and 800 pixels and randomly cropped with a probability of 0.5. For the ViT-Base model, we use Large-Scale Jittering~(LSJ)~\citep{ghiasi2021simple} with a fixed image size of 1024 following ViTDet~\citep{vitdet} and Pix2SeqV2~\citep{pix2seqv2}. The resizing range is set to [0.3, 2.0]. We essentially follow the architecture configurations of ViT-Base from ViTDet and EVA-02~\cite{eva02} with the alternated windowed attention and global attention mechanisms and extract the feature map with a stride of 32 from its simple feature pyramid network. For training time, it takes about 48 hours to train MAD-Resnet50 on 4 A100 GPUs.

\section{Autoregressive Decoding}

We convert MAD into an autoregressive variant for comparison. It follows the same architecture as well as training settings as MAD with a few modifications on tokenization of task sequences, attention mechanism, and decoding process. 

For tokenization, we adopt similar approaches as in MAD to construct task sequences for instance segmentation, keypoint detection, and image captioning, while following pix2seq~\citep{pix2seq} to build object detection sequences that the ground-truth objects are placed at the beginning of sequences. For all tasks, we add $<start>$ tokens at the start of the input sequences. During training, we add the $<end>$ tokens at the end of the original target sequences.

For the attention mechanism, the self-attention layer in the decoder is applied with a triangular causal mask for uni-directional attention.

At inference time, the task sequences are recursively generated, starting from the $<start>$ token, and generating up to the maximum length corresponding to each task~(instead of stopping at the $<end>$ token). We adopt the $argmax$ sampling strategy and cache the KV features of previous generation steps in the self-attention layers for acceleration. Although some other complex sampling strategies, i.e., beam searching or nucleus sampling~\citep{holtzman2019curious} may improve performance, these strategies would also further slow down the inference speed of autoregressive decoding.

\begin{table}[]
\footnotesize
\centering
\caption{Experimental Settings.}
    \begin{subtable}[h]{\linewidth}
    \centering
    \caption{Model with ResNet-50.}
    \begin{tabular}{l|c}
    \toprule
    config & value \\
    \midrule
    epoch & 50 \\
    optimizer & AdamW \\
    learning rate & 1e-4 \\
    learning rate scheduler & multi-step scheduler \\
    learning rate drop epoch & 40 \\
    weight decay & 1e-4 \\
    batch size & 16 \\
    image size & 800 $\times$ 1333 \\
    image augmentation & MultiScaleResize \\
    \bottomrule
    \end{tabular}
    \end{subtable}

    \begin{subtable}[h]{\linewidth}
    \centering
\vspace{10pt}
    \caption{Model with Swin-Base.}
    \begin{tabular}{l|c}
    \toprule
    config & value \\
    \midrule
    epoch & 300 \\
    optimizer & AdamW \\
    learning rate & 1e-4 \\
    learning rate scheduler & multi-step scheduler \\
    learning rate drop epoch & 240 \\
    weight decay & 1e-4 \\
    batch size & 32 \\
    image size & 800 $\times$ 1333 \\
    image augmentation & MultiScaleResize \\
    \bottomrule
    \end{tabular}
    \end{subtable}

    \begin{subtable}[h]{\linewidth}
    \centering
\vspace{7pt}
    \caption{Model with ViT-Base.}
    \begin{tabular}{l|c}
    \toprule
    config & value \\
    \midrule
    epoch & 100 \\
    optimizer & AdamW \\
    learning rate & 1e-4 \\
    learning rate scheduler & multi-step scheduler \\
    learning rate drop epoch & 80 \\
    weight decay & 1e-4 \\
    batch size & 32 \\
    image size & 1024 $\times$ 1024 \\
    image augmentation & LargeScaleJitter \\
    \bottomrule
    \end{tabular}
    \end{subtable}
\label{tab:exp_setting}
\end{table}

\section{Task Weighting}

\begin{figure*}[!t]
    \centering
    \begin{minipage}[t]{0.24\linewidth}{
        \includegraphics[width=1.0\linewidth]{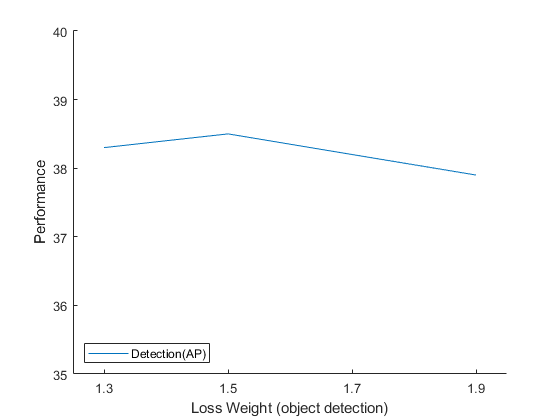}
        \subcaption{Instance segmentation Weights.}
        \label{fig:loss_det}
    }
    \end{minipage}
    \begin{minipage}[t]{0.24\linewidth}{
        \includegraphics[width=1.0\linewidth]{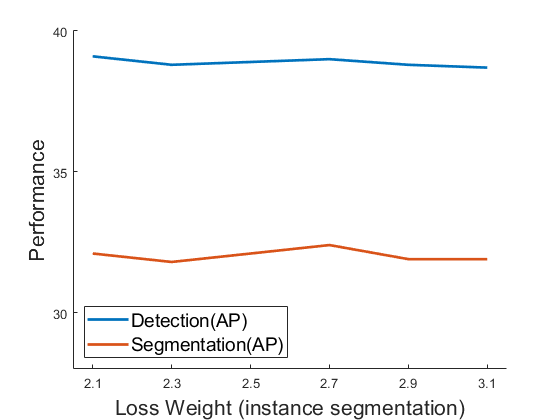}
        \subcaption{Instance segmentation Weights.}
        \label{fig:loss_seg}
    }
    \end{minipage}
    \begin{minipage}[t]{0.24\linewidth}{
        \includegraphics[width=1.0\linewidth]{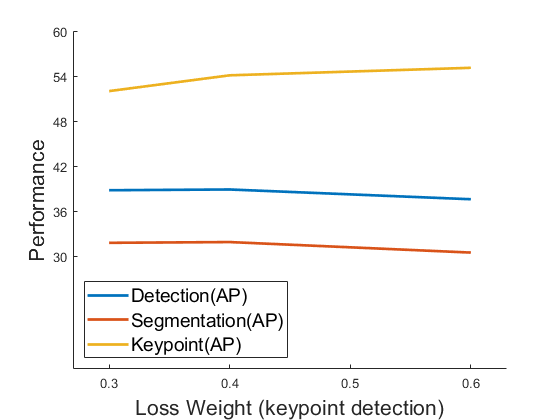}
        \subcaption{Keypoint detection weights.}
        \label{fig:loss_kpt}
    }
    \end{minipage}
    \begin{minipage}[t]{0.24\linewidth}{
        \includegraphics[width=1.0\linewidth]{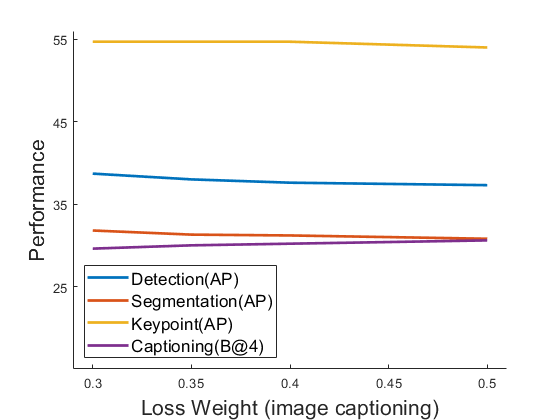}
        \subcaption{Image captioning weights.}
        \label{fig:loss_cap}
    }
    \end{minipage}
    \caption{Performance with different loss weights by gradually adding new tasks to the existing tasks. }
    \label{fig:loss_weight}
\end{figure*}

In~\cref{fig:loss_weight}, we search for the appropriate loss weight for each task. We first evaluate object detection performance and obtain the optimal loss weight of 1.5. Then we introduce the instance segmentation. As~\cref{fig:loss_seg} shows, both tasks perform well over a wide range of weights, with only small fluctuations. We thus simply take a weight of 2.7 for instance segmentation. For keypoint detection, it seems to conflict with the existing tasks, and increasing its weight would hinder the performance of object detection and instance segmentation. According to the trade-off of performance, the keypoint detection task is weighted by a factor of 0.5. Finally, we add the image captioning task, where we find that a weight of 0.3 is appropriate for preserving the performance of existing vision tasks.
